\documentclass{article}
\usepackage{times}
\usepackage{graphicx}
\usepackage{subfigure}
\usepackage{natbib}
\usepackage{algorithm}
\usepackage{algorithmic}
\usepackage{hyperref}
\usepackage{url}

\usepackage[accepted]{icml2014}
\icmltitlerunning{WordRep: A Benchmark for Research on Learning Word Representations}

\begin{document}

\twocolumn[
\icmltitle{WordRep: A Benchmark for Research on Learning Word Representations}

\icmlauthor{Bin Gao}{bingao@microsoft.com}
\icmlauthor{Jiang Bian}{jibian@microsoft.com}
\icmlauthor{Tie-Yan Liu}{tyliu@microsoft.com}
\icmladdress{Microsoft Research}

\vskip 0.3in
]

\begin{abstract}
  WordRep is a benchmark collection for the research on learning distributed word representations (or word embeddings), released by Microsoft Research. In this paper, we describe the details of the WordRep collection and show how to use it in different types of machine learning research related to word embedding. Specifically, we describe how the evaluation tasks in WordRep are selected, how the data are sampled, and how the evaluation tool is built. We then compare several state-of-the-art word representations on WordRep, report their evaluation performance, and make discussions on the results. After that, we discuss new potential research topics that can be supported by WordRep, in addition to algorithm comparison. We hope that this paper can help people gain deeper understanding of WordRep, and enable more interesting research on learning distributed word representations and related topics.
\end{abstract}

\section{Introduction}\label{sec-introduction}
The success of machine learning methods depends much on data representation, since different representations may encode different explanatory factors of variation behind the data. Conventional natural language processing (NLP) tasks often take the 1-of-$v$ word representation, where $v$ is the size of the entire vocabulary, and each word in the vocabulary is represented as a long vector with only one non-zero element. However, such simple form of word representation meets several challenges. The most critical one is that 1-of-$v$ word representation cannot indicate any relationship between different words even though they yield high semantic or syntactic correlation. For example, while \emph{elegant} and \emph{elegantly} have quite similar semantics, their corresponding 1-of-$v$ representation vectors trigger different indexes to be the hot value, and it is not explicit that \emph{elegant} is much closer to \emph{elegantly} than other words like \emph{rough} via 1-of-$v$ representations. To deal with this problem, Latent Semantic Analysis (LSA)~\cite{LSA} and Latent Dirichlet Allocation (LDA)~\cite{Blei:LDA} were proposed to learn continuous word representations. Unfortunately, it is quite difficult to train LSA or LDA model efficiently on large-scale text data.

Recently, with the rapid development of deep learning techniques, researchers have started to train complex and deep models on large amounts of text corpus, to learn distributed representations of words (also known as word embeddings) in the form of continuous vectors~\cite{Collobert2008, Bengio2003nnlm, Glorot2011, Mikolov2012phd, Socher2011RNN, TurDHH12}. While conventional NLP techniques usually represent words as indices in a vocabulary causing no notion of relationship between words, word embeddings learned by deep learning approaches aim at explicitly encoding many semantic relationships as well as linguistic regularities and patterns into the new word embedding space. For example, a previous study~\cite{Bengio2003nnlm} proposed a widely used model architecture for estimating neural network language model. Collobert et al.~\cite{Collobert2008,Collobert2011} introduced a unified neural network architecture that learns word representations based on large amounts of unlabeled training data, to deal with several different natural language processing tasks. Mikolov et al.~\cite{Mikolov2013w2v,Mikolov2013nips} proposed the continuous bag-of-words model (CBOW) and the continuous skip-gram model (Skip-gram) for learning distributed representations of words also from large amount of unlabeled text data; these two models can map the semantically or syntactically similar words to close positions in the word embedding space, based on the intuition that similar words are likely to yield similar context.

Although the study in learning distributed word representations has become very hot recently, there are very few large and public datasets for the evaluation of word representations. In this paper, we introduce a new large benchmark collection named WordRep, which is built from several different data source. We describe which evaluation tasks are selected in WordRep, how the data are sampled, and how the evaluation tool is built. We also compare the performance of several state-of-the-art word representations on WordRep. Moreover, we take further discussions on new potential research topics that can be supported by WordRep.

The rest of the paper is organized as follows. Section \ref{sec-creating} gives a detailed description about how WordRep is created. In Section \ref{sec-benchmarking}, we report the performance of several state-of-the-art word representations on WordRep. We then show how to leverage WordRep to study other research topics in Section \ref{sec-topics}. Finally, the paper is concluded in Section \ref{sec-conclusion}.

\section{Creating WordRep Collection}\label{sec-creating}
In this section, we introduce the process of creating WordRep collection, consisting of three main steps: selecting evaluation tasks, generating evaluation samples, and finalizing datasets.

\subsection{Selecting Evaluation Tasks}
The main task for WordRep is the \emph{analogical reasoning task} introduced by Mikolov et al.~\cite{Mikolov2013w2v}. The task consists of 19,544 questions, each of which is a tuple composed by two word pairs $(a, b)$ and $(c, d)$. From the tuple, the question is of the form ``$a$ is to $b$ is as $c$ is to \underline{ }\underline{ }'', denoted as $a$ : $b$ $\rightarrow$ $c$ : ?. Suppose $\overrightarrow{w}$ is the learned representation vector of word $w$ and is normalized to unit norm. Following~\cite{Mikolov2013w2v}, we answer this question by finding the word $d^*$ whose representation vector is the closest to vector $\overrightarrow{b} - \overrightarrow{a} + \overrightarrow{c}$ according to cosine similarity excluding $b$ and $c$, i.e.,
\begin{equation}
d^*=\arg\max_{x \in V, x\ne b, x\ne c}(\overrightarrow{b} - \overrightarrow{a} + \overrightarrow{c})^T\overrightarrow{x}.
\end{equation}
The question is judged as correctly-answered only if $d^*$ is exactly the answer word in the evaluation set. There are two categories of analogical tasks, including 8,869 semantic analogies in five subtasks (e.g., \emph{England} : \emph{London} $\rightarrow$ \emph{China} : \emph{Beijing}) and 10,675 syntactic analogies in nine subtasks (e.g., \emph{amazing} : \emph{amazingly} $\rightarrow$ \emph{unfortunate} : \emph{unfortunately}).

Table \ref{tab-W2V-14tasks} gives a summary of Mikolov et al.'s evaluation set. The first five subtasks are semantic questions and the rest nine subtasks are syntactic questions. The table gives one example tuple for every subtask. It also shows the number of unique word pairs and the number of tuples combined from these pairs in each subtask. Regarding to the numbers of unique word pairs, we can see that there are actually very small number of meaningful pairs in this dataset. After further checking the number of tuples, we find that not all possible combinations of word pairs are used as the tuple questions. For example, in the subtask of \emph{City-in-state}, there should be as many as 4,556 tuples combined from 68 word pairs, but only 2,467 tuples are used in the published dataset. It is not clear about the reason or how the 2,467 tuples were sampled.

\begin{table*}
\centering\vspace{-10pt}
\caption{Summary of Mikolov et al.'s evaluation set on analogical reasoning task.}
\label{tab-W2V-14tasks}
\begin{tabular}{|l||c|c||c|c||c|c|}
\hline
Subtask & \multicolumn{2}{|c||}{Word pair 1 ($a$, $b$)}  & \multicolumn{2}{|c||}{Word pair 2 ($c$, $d$)} & \# word pairs & \# tuples\\
\hline
Common capital city & Athens & Greece & Oslo & Norway & 23 & 506 \\
All capital cities & Astana & Kazakhstan & Harare & Zimbabwe & 116 & 4,524 \\
Currency & Angola & kwanza & Iran & rial & 30 & 866 \\
City-in-state & Chicago & Illinois & Stockton & California & 68 & 2,467 \\
Man-Woman & brother & sister & grandson & granddaughter & 23 & 506 \\
\hline
Adjective to adverb & apparent & apparently & rapid & rapidly & 32 & 992 \\
Opposite & possibly & impossibly & ethical & unethical & 29 & 812 \\
Comparative & great & greater & tough & tougher & 37 & 1,332 \\
Superlative & easy & easiest & lucky & luckiest & 34 & 1,122 \\
Present Participle & think & thinking & read & reading & 33 & 1,056 \\
Nationality adjective & Switzerland & Swiss & Cambodia & Cambodian & 41 & 1,599 \\
Past tense & walking & walked & swimming & swam & 40 & 1,560 \\
Plural nouns & mouse & mice & dollar & dollars & 37 & 1,332 \\
Plural verbs & work & works & speak & speaks & 30 & 870 \\
\hline
\# Total & / & / & / & / & 573 & 19,544 \\
\hline
\end{tabular}\vspace{-10pt}
\end{table*}

In the WordRep collection, we merged the above 14 subtasks into 12 subtasks, and expanded the set of word pairs in each subtask by extracting new word pairs from the Wikipedia and an English dictionary; we also added 13 new subtasks by deriving pairwise word relationship from WordNet~\cite{Wordnet}.

\subsection{Generating Evaluation Samples}
Before we describe how we generate the evaluation samples, we first introduce the scope of the vocabulary in WordRep. Since the words extracted from the Wikipedia, the dictionary, and the Web knowledge bases (WordNet) may contain many rare words that are not commonly used, we filter out those words that are not covered by the vocabulary of \emph{wiki2010}~\cite{wiki2010}. This corpus is a snapshot of the Wikipedia corpus in April 2010, which contains about two million articles and 990 million tokens. The vocabulary size of \emph{wiki2010} is 731,155, we regard which is large enough for common NLP tasks. Furthermore, note that the evaluation set proposed by Mikolov et al.~\cite{Mikolov2013w2v} contains a question set of phrase pairs besides the word pairs. To deal with phrases like \emph{New York}, they simply connect the tokens by an underline and write it as \emph{New\_York}. In this release of WordRep, we will only generate question sets with word pairs and leave phrase pairs for the future versions.

\subsubsection{Wikipedia Knowledge}
We leverage Wikipedia knowledge to enlarge the semantic analogical tasks. For the first two subtasks in Table \ref{tab-W2V-14tasks}, we merged \emph{Common capital city} into \emph{All capital cities} for simplicity. Then, we extracted the Wikipedia pages for all the counties and areas to get a full list of countries, capitals, currencies, and nationality adjectives, so that we can enlarge the question sets in the subtasks of \emph{All capital cities}, \emph{Currency}, and \emph{Nationality adjective}. For \emph{City-in-state}, we found the top cities in population in the 50 states in the U.S. from Wikipedia, and built the city-state pairs accordingly. Note that we only kept the word based name entities, and removed the phrase based names.

\subsubsection{Dictionary Knowledge}
We take advantage of dictionary knowledge to enlarge both semantic and syntactic analogical tasks. For the subtask \emph{Opposite}, we merged it into a new subtask \emph{Antonym} extracted from WordNet, which will be described in the following. For the rest subtasks (\emph{Man-Woman}, \emph{Adjective to adverb}, \emph{Comparative}, \emph{Superlative}, \emph{Present Participle}, \emph{Past tense}, \emph{Plural nouns}, \emph{Plural verbs}) in Table \ref{tab-W2V-14tasks}, we enlarged their corresponding word pairs by extracting new candidates from the Longman Dictionaries\footnote{\url{http://www.longmandictionariesonline.com/}}. For \emph{Man-Woman}, we also added some word pairs of male and female animals like \emph{cock} and \emph{hen}. Note that the newly extracted words are filtered by the vocabulary of \emph{wiki2010}. The statistics about the enlarged subtasks by Wikipedia knowledge and Dictionary knowledge is shown in Table \ref{tab-enlarged-12tasks}.

\subsubsection{WordNet Knowledge}
WordNet~\cite{Wordnet} is a large lexical database of English. It contains both the words and the relations among the words, which is a gold mine to enlarge both semantic and syntactic analogical tasks. We extracted 13 new subtasks based on the WordNet relations including \emph{Antonym}, \emph{MemberOf}, \emph{MadeOf}, \emph{IsA}, \emph{SimilarTo}, \emph{PartOf}, \emph{InstanceOf}, \emph{DerivedFrom}, \emph{HasContext}, \emph{RelatedTo}, \emph{Attribute}, \emph{Causes}, and \emph{Entails}. Note that we merged the original subtask \emph{Opposite} in Table \ref{tab-W2V-14tasks} into the new subtask \emph{Antonym}, and we also filtered the words in the 13 new subtasks by the vocabulary of \emph{wiki2010}. The statistics and examples about the 13 WordNet subtasks are shown in Table \ref{tab-wordnet-13tasks}.

\subsection{Finalizing Datasets}
From Table \ref{tab-enlarged-12tasks} and Table \ref{tab-wordnet-13tasks}, we can see that WordRep has much larger number of word pairs as well as word tuples. The word pairs and word tuples can be downloaded ({\em http://research.microsoft.com/en-us/um/beijing/events/kpdltm2014/WordRep-1.0.zip}). The size of the compressed package is 1.61 GB.

\section{Evaluation on The Benchmark Collection}\label{sec-benchmarking}
In this section, we report the performance of several state-of-the-art distributed word representations on WordRep, including CW08~\cite{Collobert2008}, RNNLM~\cite{Mikolov2012phd}, and CBOW~\cite{Mikolov2013w2v}. In particular, we download the public word representations of CW08\footnote{\url{http://ml.nec-labs.com/senna/}}~\cite{Collobert2008} whose dimension is 50; the public word representations of RNNLM~\cite{Mikolov2012phd} are obtained from its public site\footnote{\url{http://www.fit.vutbr.cz/~imikolov/rnnlm/}}, which includes three word representation models with the dimension of 80, 640, and 1600, respectively; we obtain the CBOW~\cite{Mikolov2013w2v} models by using its online tools\footnote{\url{https://code.google.com/p/word2vec/}} to train word representations directly on the {\em wiki2010} dataset, where we set the dimension as 100, 200, and 300, respectively.

Table~\ref{tab-perf-enlarged-12tasks} demonstrates the accuracy of each of word representations on the enlarged evaluation set of analogical reasoning tasks. From this table, we can find that different word representations yield quite various accuracy on the analogical reasoning tasks. In particular, CW08 with 50 dimension can only achieve relatively low accuracy compared with RNNLM and CBOW, which may be due to the difference in terms of the dimension of word representations, the training data, or the training algorithms. We can also find that, with respect to the same training method, larger dimension of word representations is more likely to result in better performance.

Table~\ref{tab-perf-wordnet-13tasks} demonstrates the accuracy of each of word representations on the WordNet evaluation set of analogical reasoning tasks. From this table, we can see the analogous observations with the Table~\ref{tab-perf-enlarged-12tasks}. From these two tables, we can also find that different subtasks could yield quite diverse accuracy by evaluating on the same word representation model.

Note that the above evaluation experiments can be done using the evaluation tool provided by Word2Vec\footnote{\url{https://code.google.com/p/word2vec/}}. The only difference is that we treat all tuple questions as \emph{seen} questions. Thus, if a tuple question is \emph{unseen} by the word embedding being evaluated (i.e., at least one word in the tuple question is not in the vocabulary of the word embedding), we will regard it to be answered incorrectly.

\section{Supporting New Research Topics}\label{sec-topics}
Besides measuring the quality of word representations, we can also use WordRep for, but not limited to, the following tasks.
\begin{itemize}
  \item WordRep can be used to evaluate the embedding for relations. Recently some researchers have attempted to do word embedding and relation embedding simultaneously. WordRep contains 25 subtasks in total, i.e., there are 25 types of relations that can be used for evaluating relation embeddings.
  \item WordRep can be used to evaluate relation prediction. The 25 types of relations can be regarded as labels of word pairs. Researchers can test their elation prediction methods using these labels as ground-truth.
  \item WordRep provides several good word lists for general NLP tasks. For example, there are lists for different syntactic forms of nouns, verbs, adjectives, and adverbs, and there are lists for commonly used relations.
\end{itemize}

\section{Conclusions}\label{sec-conclusion}
In this paper, we have introduced a new data collection called WordRep, which can be used for the evaluation of distributed word representations. We described how we built the data collection and reported the evaluation performance of several state-of-the-art word representations on it. We also discussed the possible research topics that WordRep may support.

For the future work, we plan to further expand the evaluation set to phrase pairs, and we also plan to enrich the collection by considering other Web knowledge bases like Freebase~\cite{freebase}.

\section{Acknowledgements}
We would like to thank Siyu Qiu, Chang Xu, Yalong Bai, Hongfei Xue, and Rui Zhang for their contributions in the preparation of the collection and the evaluation of the state-of-the-art word embeddings.

\begin{table*}
\centering\vspace{0pt}
\caption{Summary of enlarged evaluation set on analogical reasoning task.}
\label{tab-enlarged-12tasks}
\begin{tabular}{|l||c|c||c|c||c|c|}
\hline
Subtask & \multicolumn{2}{|c||}{Word pair 1 ($a$, $b$)}  & \multicolumn{2}{|c||}{Word pair 2 ($c$, $d$)} & \# word pairs & \# tuples\\
\hline
All capital cities & Astana & Kazakhstan & Harare & Zimbabwe & 131 & 17,030 \\
Currency & Angola & kwanza & Iran & rial & 119 & 14,042 \\
City-in-state & Chicago & Illinois & Stockton & California & 3,296 & 10,860,320 \\
Man-Woman & brother & sister & grandson & granddaughter & 90 & 8,010 \\
\hline
Adjective to adverb & apparent & apparently & rapid & rapidly & 689 & 474,032 \\
Comparative & great & greater & tough & tougher & 97 & 9,312 \\
Superlative & easy & easiest & lucky & luckiest & 87 & 7,482 \\
Present Participle & think & thinking & read & reading & 2,951 & 8,705,450 \\
Nationality adjective & Switzerland & Swiss & Cambodia & Cambodian & 139 & 19,182 \\
Past tense & walking & walked & swimming & swam & 2,827 & 7,989,102 \\
Plural nouns & mouse & mice & dollar & dollars & 5,868 & 34,427,556 \\
Plural verbs & work & works & speak & speaks & 2,708 & 7,330,556 \\
\hline
\# Total & / & / & / & / & 19,002 & 69,862,074 \\
\hline
\end{tabular}
\end{table*}

\begin{table*}
\centering\vspace{0pt}
\caption{Summary of WordNet evaluation set on analogical reasoning task.}
\label{tab-wordnet-13tasks}
\begin{tabular}{|l||c|c||c|c||c|c|}
\hline
Subtask & \multicolumn{2}{|c||}{Word pair 1 ($a$, $b$)}  & \multicolumn{2}{|c||}{Word pair 2 ($c$, $d$)} & \# word pairs & \# tuples\\
\hline
Antonym & succeed & fail & major & minor & 973 & 945,756 \\
MemberOf & Zealander & Zealand & Malawian & Malawi & 406 & 164,430 \\
MadeOf & tear & water & glassware & glass & 63 & 3,906 \\
IsA & yoga & exercise & yogurt & food & 10,615 & 112,667,610 \\
SimilarTo & worthy & value & employ & work & 3,489 & 12,169,632 \\
PartOf & eyebrow & face & column & temple & 1,029 & 1,057,812 \\
InstanceOf & Dallas & city & Somerset & county & 1,314 & 1,725,282 \\
DerivedFrom & king & kingdom & liberate & liberty & 6,119 & 37,436,042 \\
HasContext & underhand & sport & rhyme & poetry & 1,149 & 1,319,052 \\
RelatedTo & look & watch & use & consume & 102 & 10,302 \\
Attribute & size & small & depth & shallow & 184 & 33,672 \\
Causes & ring & sound & give & have & 26 & 650 \\
Entails & buy & pay & beat & win & 114 & 12,882 \\
\hline
\# Total & / & / & / & / & 25,583 & 167,547,028 \\
\hline
\end{tabular}
\end{table*}

\begin{table*}
\centering\vspace{0pt}
\caption{Accuracy of various word representations on the enlarged evaluation set of analogical reasoning tasks.}
\label{tab-perf-enlarged-12tasks}
\begin{tabular}{|l||c||c|c|c||c|c|c||}
\hline
 & CW08 & \multicolumn{3}{|c||}{RNNLM} & \multicolumn{3}{|c||}{CBOW} \\
\hline
Subtask & dim=50 & dim=80 & dim=640 & dim=1600 & dim=100 & dim=200 & dim=300 \\
\hline
All capital cities & 0.62\% & 0.76\% & 1.23\% & 1.81\% & 6.62\% & 9.04\% & 11.28\% \\
Currency & 0.25\% & 0.41\% & 0.66\% & 0.87\% & 3.13\% & 4.06\% & 4.32\% \\
City-in-state & 0.67\% & 1.36\% & 3.14\% & 3.38\% & 1.55\% & 1.86\% & 2.25\% \\
Man-Woman & 4.83\% & 6.67\% & 18.46\% & 20.82\% & 25.89\% & 29.06\% & 28.60\% \\
\hline
Adjective to adverb & 1.40\% & 0.93\% & 1.17\% & 2.01\% & 3.45\% & 3.44\% & 3.23\% \\
Comparative & 1.55\% & 16.61\% & 34.92\% & 40.28\% & 33.41\% & 41.70\% & 42.53\% \\
Superlative & 1.94\% & 9.18\% & 25.33\% & 26.21\% & 23.56\% & 24.99\% & 29.07\% \\
Present participle & 1.53\% & 9.67\% & 20.03\% & 23.26\% & 8.20\% & 11.25\% & 11.75\% \\
Nationality adjective & 3.07\% & 1.62\% & 3.15\% & 3.76\% & 23.66\% & 40.19\% & 47.44\% \\
Past tense & 1.84\% & 10.43\% & 19.51\% & 22.77\% & 15.51\% & 21.60\% & 24.15\% \\
Plural nouns & 3.21\% & 3.32\% & 14.42\% & 18.28\% & 23.95\% & 34.64\% & 38.82\% \\
Plural verbs & 2.44\% & 7.89\% & 22.41\% & 26.62\% & 17.28\% & 27.47\% & 31.82\% \\
\hline
Total & 2.36\% & 5.08\% & 14.69\% & 17.85\% & 16.70\% & 24.16\% & 27.10\% \\
\hline
\end{tabular}
\end{table*}

\begin{table*}
\centering\vspace{0pt}
\caption{Accuracy of various word representations on the WordNet evaluation set of analogical reasoning tasks.}
\label{tab-perf-wordnet-13tasks}
\begin{tabular}{|l||c||c|c|c||c|c|c||}
\hline
 & CW08 & \multicolumn{3}{|c||}{RNNLM} & \multicolumn{3}{|c||}{CBOW} \\
\hline
Subtask & dim=50 & dim=80 & dim=640 & dim=1600 & dim=100 & dim=200 & dim=300 \\
\hline
Antonym & 0.28\% & 1.21\% & 2.88\% & 3.12\% & 2.74\% & 4.08\% & 4.57\% \\
Attribute & 0.22\% & 0.11\% & 0.24\% & 0.42\% & 0.68\% & 1.09\% & 1.18\% \\
Causes & 0.00\% & 0.31\% & 0.00\% & 0.00\% & 0.15\% & 0.31\% & 1.08\% \\
DerivedFrom & 0.05\% & 0.06\% & 0.16\% & 0.18\% & 0.33\% & 0.53\% & 0.63\% \\
Entails & 0.05\% & 0.05\% & 0.05\% & 0.07\% & 0.26\% & 0.44\% & 0.38\% \\
HasContext & 0.12\% & 0.06\% & 0.16\% & 0.19\% & 0.28\% & 0.38\% & 0.35\% \\
InstanceOf & 0.08\% & 0.24\% & 0.81\% & 0.64\% & 0.48\% & 0.60\% & 0.58\% \\
IsA & 0.07\% & 0.18\% & 0.42\% & 0.47\% & 0.42\% & 0.64\% & 0.67\% \\
MadeOf & 0.03\% & 0.15\% & 0.10\% & 0.13\% & 0.33\% & 1.02\% & 0.72\% \\
MemberOf & 0.08\% & 0.07\% & 0.11\% & 0.13\% & 0.58\% & 0.84\% & 1.06\% \\
PartOf & 0.31\% & 0.29\% & 0.55\% & 0.60\% & 1.17\% & 1.24\% & 1.27\% \\
RelatedTo & 0.00\% & 0.04\% & 0.02\% & 0.00\% & 0.20\% & 0.09\% & 0.05\% \\
SimilarTo & 0.02\% & 0.07\% & 0.14\% & 0.18\% & 0.14\% & 0.23\% & 0.29\% \\
\hline
Total & 0.06\% & 0.15\% & 0.35\% & 0.40\% & 0.40\% & 0.61\% & 0.66\% \\
\hline
\end{tabular}
\end{table*}

\bibliography{wordrep}

\begin{thebibliography}{14}
\providecommand{\natexlab}[1]{#1}
\providecommand{\url}[1]{\texttt{#1}}
\expandafter\ifx\csname urlstyle\endcsname\relax
  \providecommand{\doi}[1]{doi: #1}\else
  \providecommand{\doi}{doi: \begingroup \urlstyle{rm}\Url}\fi

\bibitem[Bengio et~al.(2003)Bengio, Ducharme, Vincent, and
  Janvin]{Bengio2003nnlm}
Bengio, Y., Ducharme, R., Vincent, P., and Janvin, C.
\newblock A neural probabilistic language model.
\newblock In \emph{The Journal of Machine Learning Research}, pp.\
  3:1137--1155, 2003.

\bibitem[Blei et~al.(2003)Blei, Ng, and Jordan]{Blei:LDA}
Blei, David~M., Ng, Andrew~Y., and Jordan, Michael~I.
\newblock Latent dirichlet allocation.
\newblock \emph{J. Mach. Learn. Res.}, 3:\penalty0 993--1022, March 2003.
\newblock ISSN 1532-4435.

\bibitem[Bollacker et~al.(2008)Bollacker, Evans, Paritosh, Sturge, and
  Taylor]{freebase}
Bollacker, K., Evans, C., Paritosh, P., Sturge, T., and Taylor, J.
\newblock Freebase: a collaboratively created graph database for structuring
  human knowledge.
\newblock In \emph{Proceedings of the 2008 ACM SIGMOD international conference
  on Management of data}, pp.\  1247--1250. ACM, 2008.

\bibitem[Collobert \& Weston(2008)Collobert and Weston]{Collobert2008}
Collobert, R. and Weston, J.
\newblock A unified architecture for natural language processing: Deep neural
  networks with multitask learning.
\newblock In \emph{Proceedings of the 25th International Conference on Machine
  Learning}, ICML '08, pp.\  160--167, New York, NY, USA, 2008. ACM.

\bibitem[Collobert et~al.(2011)Collobert, Weston, Bottou, Karlen, Kavukcuoglu,
  and Kuksa]{Collobert2011}
Collobert, R., Weston, J., Bottou, L., Karlen, M., Kavukcuoglu, K., and Kuksa,
  P.
\newblock Natural language processing (almost) from scratch.
\newblock \emph{Journal of Machine Learning Research}, 12:\penalty0 2493--2537,
  November 2011.
\newblock ISSN 1532-4435.

\bibitem[Dumais()]{LSA}
Dumais, Susan~T.
\newblock Latent semantic analysis.
\newblock \emph{Annual Review of Information Science and Technology},
  38\penalty0 (1).
\newblock ISSN 1550-8382.

\bibitem[Glorot et~al.(2011)Glorot, Bordes, and Bengio]{Glorot2011}
Glorot, X., Bordes, A., and Bengio, Y.
\newblock Domain adaptation for large-scale sentiment classification: A deep
  learning approach.
\newblock In \emph{In Proceedings of the Twenty-eight International Conference
  on Machine Learning, ICML}, 2011.

\bibitem[Mikolov(2012)]{Mikolov2012phd}
Mikolov, T.
\newblock \emph{Statistical Language Models Based on Neural Networks}.
\newblock PhD thesis, Brno University of Technology, 2012.

\bibitem[Mikolov et~al.(2013{\natexlab{a}})Mikolov, Chen, Corrado, and
  Dean]{Mikolov2013w2v}
Mikolov, T., Chen, K., Corrado, G., and Dean, J.
\newblock Efficient estimation of word representations in vector space.
\newblock \emph{CoRR}, abs/1301.3781, 2013{\natexlab{a}}.

\bibitem[Mikolov et~al.(2013{\natexlab{b}})Mikolov, Sutskever, Chen, Corrado,
  and Dean]{Mikolov2013nips}
Mikolov, T., Sutskever, I., Chen, K., Corrado, G.~S., and Dean, J.
\newblock Distributed representations of words and phrases and their
  compositionality.
\newblock In Burges, Christopher J.~C., Bottou, Léon, Ghahramani, Zoubin, and
  Weinberger, Kilian~Q. (eds.), \emph{NIPS}, pp.\  3111--3119,
  2013{\natexlab{b}}.

\bibitem[Shaoul \& Westbury(2010)Shaoul and Westbury]{wiki2010}
Shaoul, C. and Westbury, C.
\newblock The westbury lab wikipedia corpus, edmonton, ab: University of
  alberta, 2010.

\bibitem[Socher et~al.(2011)Socher, Lin, Ng, and Manning]{Socher2011RNN}
Socher, R., Lin, C.~C., Ng, A.~Y., and Manning, C.~D.
\newblock Parsing natural scenes and natural language with recursive neural
  networks.
\newblock In \emph{Proceedings of the 26th International Conference on Machine
  Learning (ICML)}, 2011.

\bibitem[Tur et~al.(2012)Tur, Deng, Hakkani-Tur, and He]{TurDHH12}
Tur, G., Deng, L., Hakkani-Tur, D., and He, X.
\newblock Towards deeper understanding: Deep convex networks for semantic
  utterance classification.
\newblock In \emph{ICASSP}, pp.\  5045--5048, 2012.

\bibitem[WordNet(2010)]{Wordnet}
WordNet.
\newblock ``about wordnet'', princeton university, 2010.
\newblock URL \url{http://wordnet.princeton.edu}.

\end{thebibliography}
\bibliographystyle{icml2014}

\end{document}